\documentclass[sigconf, noacm]{acmart}

\usepackage{multirow}
\usepackage{multicol}
\usepackage{subcaption}

\AtBeginDocument{%
  \providecommand\BibTeX{{%
    \normalfont B\kern-0.5em{\scshape i\kern-0.25em b}\kern-0.8em\TeX}}}

\setcopyright{rightsretained}
\acmConference[]{}{}{}

\begin{document}

\title[Empty-shelf Detection]{Designing an Efficient End-to-end Machine Learning Pipeline for Real-time Empty-shelf Detection}

\author{Dipendra Jha}
\email{dipendra.jha@target.com}
\orcid{0000-0002-6210-1937}
\affiliation{%
  \institution{Target Corporation}
  \country{USA}
}

\author{Ata Mahjoubfar}
\email{ata.mahjoubfar@target.com}
\orcid{0000-0001-5702-6760}
\affiliation{%
  \institution{Target Corporation}
  \country{USA}
}

\author{Anupama Joshi}
\email{anupama.joshi@target.com}
\affiliation{%
  \institution{Target Coorporation}
  \country{USA}
}

\renewcommand{\shortauthors}{Jha et al.}

\begin{abstract}
  On-Shelf Availability (OSA) of products in retail stores is a critical business criterion in the fast moving consumer goods and retails sector. When a product is out-of-stock (OOS) and a customer cannot find it on its designed shelf, this motivates the customer to store-switching or buying nothing, which causes fall in future sales and demands. Retailers are employing several approaches to detect empty shelves and ensure high OSA of products; however, such methods are generally ineffective and infeasible since they are either manual, expensive or less accurate. Recently machine learning based solutions have been proposed, but they suffer from high computational cost and low accuracy problem due to lack of large annotated datasets of on-shelf products. Here, we present an elegant approach for designing an end-to-end machine learning (ML) pipeline for real-time empty shelf detection. Considering the strong dependency between the quality of ML models and the quality of data, we focus on the importance of proper data collection, cleaning and correct data annotation before delving into modeling. Since an empty-shelf detection solution should be computationally-efficient for real-time predictions, we explore different run-time optimizations to improve the model performance. Our dataset contains 1000 images, collected and annotated by following well-defined guidelines. Our low-latency model achieves a mean average F1-score of 68.5\%, and can process up to 67 images/s on Intel Xeon Gold and up to 860 images/s on an A100 GPU.
\end{abstract}


\maketitle

\section{Introduction}

\begin{figure*}[!t]
  \includegraphics[width=0.99\textwidth, keepaspectratio]{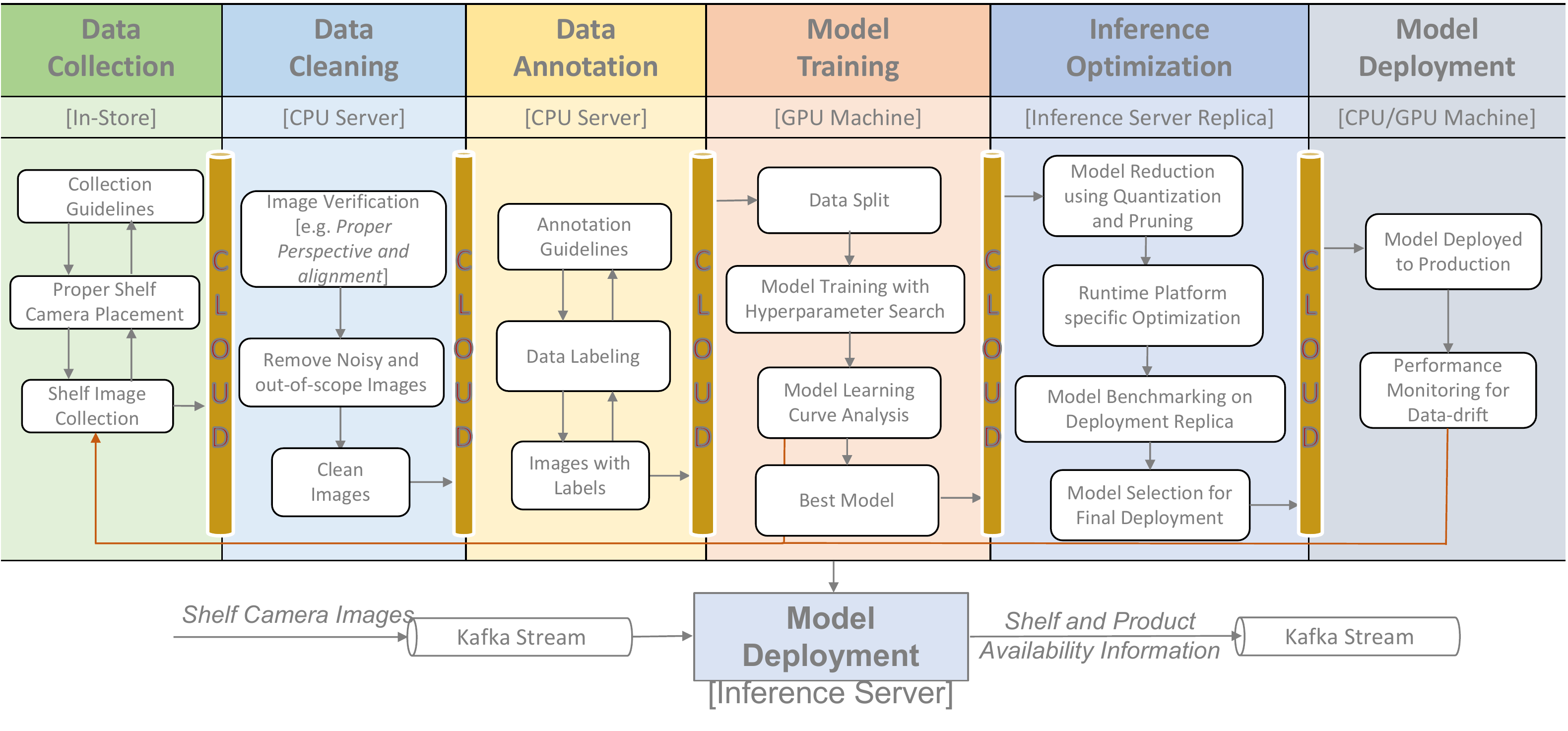}
  \caption{End-to-end machine learning pipeline for real-time empty-shelf detection.}
  \Description{}
  \label{fig:pipeline}
\end{figure*}

Retail operation is all about offering customers access to their wanted shopping items at the appropriate shelf locations when they visit the store~\cite{fisher2000rocket}. On-Shelf-Availability (OSA) of products has been deemed a critical measure of a successful retail business operation due to its impact on the current and future demand~\cite{anderson2006measuring, gruen2007comprehensive}. Out-of-Stock (OOS) occurs when a customer at a retail store wants to buy  a product that is not currently available at its designated shelf~\cite{spielmaker2012shelf}. Musalem et al.~\cite{musalem2010structural} reports that the OOS rate is significantly high in the United States and Europe; the costs associated to OOS problem vary across product categories and can be substantial in some cases~\cite{gruen2002retail}.
Whenever the OOS issue repeats continuously, the customer go to a another store with OSA of their needed products, considering the wide range of retail stores available today. 
A significant portion of these issues remain unresolved for significant duration of time~\cite{mitchell2012improving}; this has a harsh impact on the performance and profitability of the retail store. Improvement in OSA has greater impact on sales for a retailer than a manufacturer~\cite{mitchell2012improving}. 
Therefore, retail stores have to  ensure high OSA to retain their long-term permanent customers. OOS can have significant impact on the business profit and it has become an integrated measure of retailer's performance given its labor, processes and technology.

Retailers employ various approaches to mitigate the Out-of-shelf challenges and maximize product On-shelf availability.
Such methods range from  manual store audits, leveraging of Radio Frequency Identification (RFID) and RFID reader integrated weight sensing mat to ZigBee transceiver and consumer-grade depth sensor to customer-centric approach of scanning the QR code and alerting the store manager~\cite{chao2007determining, milella20213d, moorthy2015applying}; such methods are either manual or not cost-effective to integrate into existing systems~\cite{michael2005pros}. 
There have also been proposal to address the OOS issue using image processing and traditional machine learning algorithms such as using image processing~\cite{moorthy2015applying}, supervised learning using Support Vector Machines (SVM)~\cite{rosado2016supervised}, blob detection followed by discriminative machine learning~\cite{muthugnanambika2018automated}, 3D point cloud reconstruction and modeling~\cite{milella2020towards}, and computer vision~\cite{priyanwada2020benchmark}.
Nevertheless, such traditional ML approaches have low accuracy even using larger datasets and are difficult to make better.
To address the low accuracy of such methods, there exist some recent work leveraging deep learning approaches to address the issue of shelf-OOS~\cite{higa2019robust, chen2019out, rong2020solution, yilmazer2021shelf}. Some recent works have focused on product detection to demonstrate object detection in densely packed scenes~\cite{goldman2019precise, rong2020solution, varadarajan2020benchmark}.
When a deep learning (DL) approach is used, comparatively higher accuracies have been achieved.
However, since DL methods require labelled samples for training, a huge manual effort is required to properly annotate products on retail shelves for building a DL-based predictive model~\cite{wei2020deep}.

The main challenge behind the application of deep learning model in retail field remains to be accessibility and availability of good quality data.
A predictive application based on machine learning (ML) is a software artifact compiled from data~\cite{karpathy2017software}.
The biggest perceived problem in MLOps revolves around data collection and preprocessing~\cite{makinen2021needs}; collecting, cleaning, annotating and preprocessing of real world data can become as challenging and complex as training, deployment and monitoring of ML models. %~\cite{John1586498}.
The performance of an ML model is strongly dependent on the quality of dataset used for training the model~\cite{renggli2021data}. Renggli et al.~\cite{renggli2021data} splits the quality of data across four dimensions of accuracy, completeness, consistency and timeliness. The accuracy of dataset deals with correctness and reliability, the completeness stands for inclusivity of real world scenarios, consistency deals with rules followed for data collection and preparation, and timeliness deals with whether the data is up-to-date for the task~\cite{renggli2021data}.
Although, there are some datasets collected in retail stores, such as SKU-110k~\cite{goldman2019precise} and WebMarket~\cite{webmarket}, they lack data quality measured across all four dimensions.

In this paper, we present an elegant approach for designing an end-to-end machine learning (ML) pipeline for real-time empty-shelf detection to address the OOS problem and ensure high OSA (Figure~\ref{fig:pipeline}). Considering the strong dependency between the quality of ML models and the quality of data, we focus the first three stage of our ML pipeline on improving the data quality, before delving into modeling. Existing DL works have focused on product detection to solve the shelf-OOS issue~\cite{higa2019robust, chen2019out, rong2020solution, yilmazer2021shelf, goldman2019precise, rong2020solution, varadarajan2020benchmark}.
Here, we focus entirely on empty-shelf detection rather than product detection; our approach tremendously decreases human efforts needed for data annotations as well as model computation time; an alternate analogy would be to focus on detecting the empty road space to drive rather than focusing on the entire road environment when building an AI-based self-driving solution. To make our solution efficient for real-time predictions, we use state-of-the-art real-time object detection model architectures, followed by inference run-time optimizations for different computing devices to improve performance. Our dataset contains 1000 images, collected and annotated by following proper well-defined guidelines; the optimized model with lowest latency achieves a mean average F1-score of 68.5\% on our test set, and can process up to 67 images/s on Intel Xeon Gold and up to around 860 images/s on an A100 GPU. 
The rest of the paper discusses the background and related work, followed by six stages of ML pipeline in order; finally we discuss the significance of the work and conclude.

\section{Background and Related Works}
Retail Out-of-stock (OOS) can be classified into two subcategories -- store-OOS, where the item is not currently available; or shelf-OOS where the item is in store but customer can not find it since the item is not put not in the correct location. When an in-store OOS condition occurs, customer reaction varies from store switching, brand switching to even not buying anything, which leads to huge loss in sales and revenue~\cite{corsten2003desperately, mitchell2012improving, hausruckinger2006approaches}.
OSA is checked by employees manually at most retail stores; this method are generally ineffective and unsustainable because it requires continuous manual effort.

Several image processing and machine learning based solutions have been proposed for the OOS problem. 
Milella et al.~\cite{milella20213d} presented a 3D Vision-Based Shelf Monitoring System (3D-VSM) aimed at automatically estimating the OSA of products in retail stores by exploiting 3D data returned by a consumer-grade depth sensor. 
Moorthy et al.~\cite{moorthy2015applying} proposed image processing techniques for detecting the front-facing products as well as the empty shelf locations.
Rosado et al.~\cite{rosado2016supervised} proposed a supervised machine learning method leveraging Support Vector Machines for OOS detection by leveraging the geometrical and visual features in a high-resolution panoramic shelf images of grocery retail stores.
Milella et al.~\cite{milella2020towards} developed an early detection method for OOS situations based on low-cost embedded system by exploiting 3D point cloud reconstruction and modeling techniques.
Pranwada et al.~\cite{priyanwada2020benchmark} presented a computer vision and machine learning based approach for detecting empty shelves from camera images. 
Recently, there have been some works based on deep learning approaches to address the OOS issue.
Higa et al.~\cite{higa2019robust} proposed an approach for monitoring product shelves robustly by classifying the detected change regions into ``product take" vs ``product replenished/returned" using deep learning. 
Chen et al.~\cite{chen2019out} used Faster R-CNN algorithm to obtain location information, followed by several OOS detection approaches.
Goldman et al.~\cite{goldman2019precise} presented an annotated dataset- SKU-110k, which contains 11,762 images with 110k classes, representing the retail environments with densely packed scenes containing numerous objects placed in close proximity; 
they applied a variant of RetinaNet with their EM-Merger head and Soft-IoU and achieved an average precision of 49.2\%. 
Rong et al.~\cite{rong2020solution} proposed a variant of random cropping strategy with optimized Cascade R-CNN to solve the densely packed scene detection of retail stores and achieved a mean average precision of 58.7\% on SKU-110k~\cite{goldman2019precise}.
Varadarajan et al.~\cite{varadarajan2020benchmark} showed that they could achieve satisfactory results (mAP=0.56) at low IoU of (0.5) using RetinaNet using only 312 images. 
Wei et al.~\cite{wei2020deep} presented a comprehensive deep learning research review aimed at retail product recognition.
Recently, Yilmazer and Birant~\cite{yilmazer2021shelf} used 1500 images from WebMarket dataset~\cite{webmarket} and used YOLOv4 for empty, almost empty and three product class detection; they iteratively trained their models using pseudo-labelled data labelled using their best model from last iteration.

\section{Data Collection}

We collected shelf images from Target retail stores at different locations in the United States. There can be multiple issues with the captured images if proper guidelines are not followed for camera placement and camera settings. For example, some cameras can have fish-eye distortions, variations in positioning of the lenses, and various zoom levels; these distortions can not be compensated using planar projections. In this study, we experimented with collecting images using a personal mobile camera (iPhone 12 Pro Max). We only focused on collecting images from the paper towels and tissue paper aisles since these products have well-defined rectangular shapes with vertical boundaries; this makes the marking of the bounding boxes for empty locations easier. To ensure all four dimensions of data quality~\cite{batini2009methodologies}, we followed a well-defined data collection guideline:

\begin{enumerate}
    \item Align the camera position parallel to the shelf so that the image borders are parallel to the shelf. If all borders can not be properly aligned, try to align the left and bottom sides [accuracy, consistency].
    
    \item {Fix the camera position and focal length so that it captures full shelf height, from top to bottom. [accuracy, consistency]}
    
    \item {Collect shelf images from a different time of the day and other weekdays such that the dataset contains different distributions of empty-shelve counts. Shelves are generally filled during the early morning and get empty during the late evening [timeliness, completeness].}
    
    \item {Collect multiple shelf images for the same shelf positions at different times such that the dataset captures different products versus empty representations for the same shelf [timeliness, completeness].}
\end{enumerate}

Note that different shelves can have different lengths and heights; while some shelves could be completely covered by 3 consecutive image frames, others required more image frames. Following our guidelines, we incrementally collected over 1080 images in this study from different locations across United States in 2021.

\section{Data Cleaning}
Following a proper well-defined data collection guideline is important for building a real-world machine learning solution; it can not only avoid noisy image collection, but also helps in reducing human efforts required for further data collection and data cleaning. However, since the data collection was performed manually in a real retail environment, the collected data still contained some images that narrowly missed the data collection guidelines. Some examples in our case include images with poor border alignments, images containing irregular product shapes, images not having enough top view, images with people shopping, and so on. Note that we limit ourselves to the main aisle face of the shelves in this study; we do not collect images or label the side faces. The use of data collection guidelines helped us in tremendously reducing the noisy images.

\begin{figure}[!b]
  \includegraphics[width=0.4\textwidth, keepaspectratio]{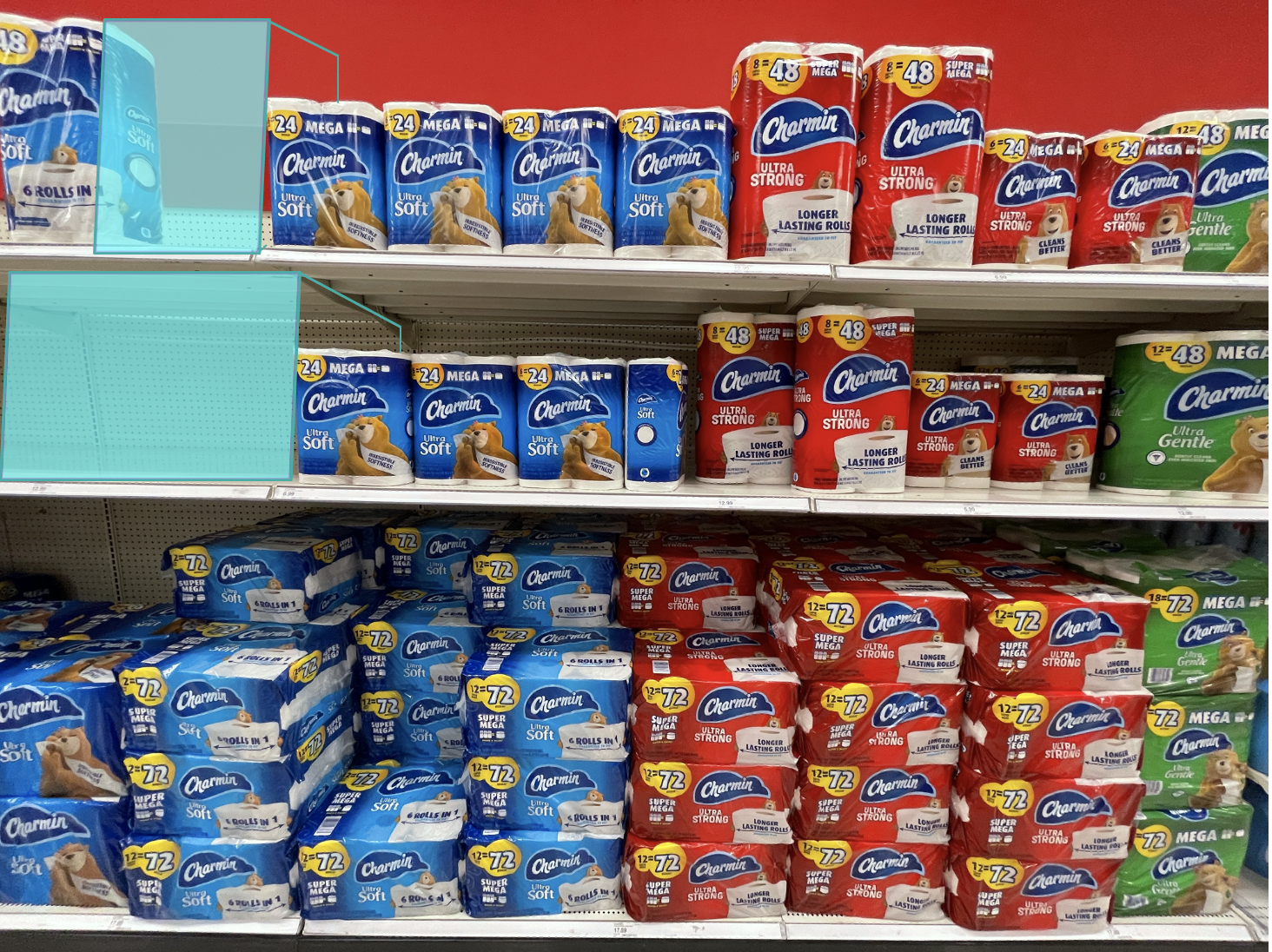}
  \caption{Labeling the front face of empty location visualized as a 3D box.}
  \Description{}
  \label{fig:labeling_3D}
\end{figure}

\begin{figure}[!b]
  \includegraphics[width=0.45\textwidth, keepaspectratio]{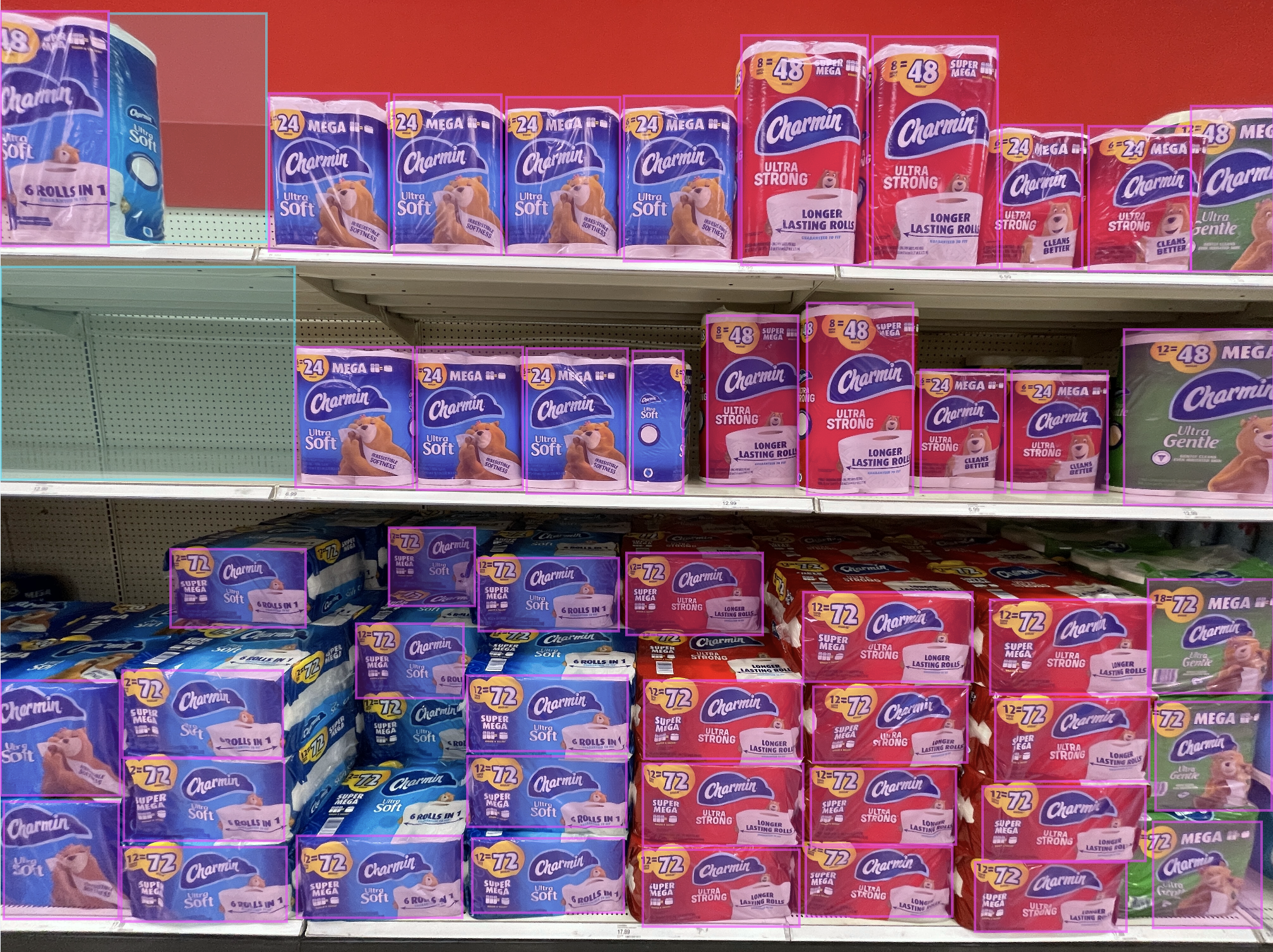}
  \caption{Empty locations and the products.}
\label{fig:labeling}
\end{figure}

\begin{figure*}[!tb]
  \includegraphics[width=0.9\textwidth]{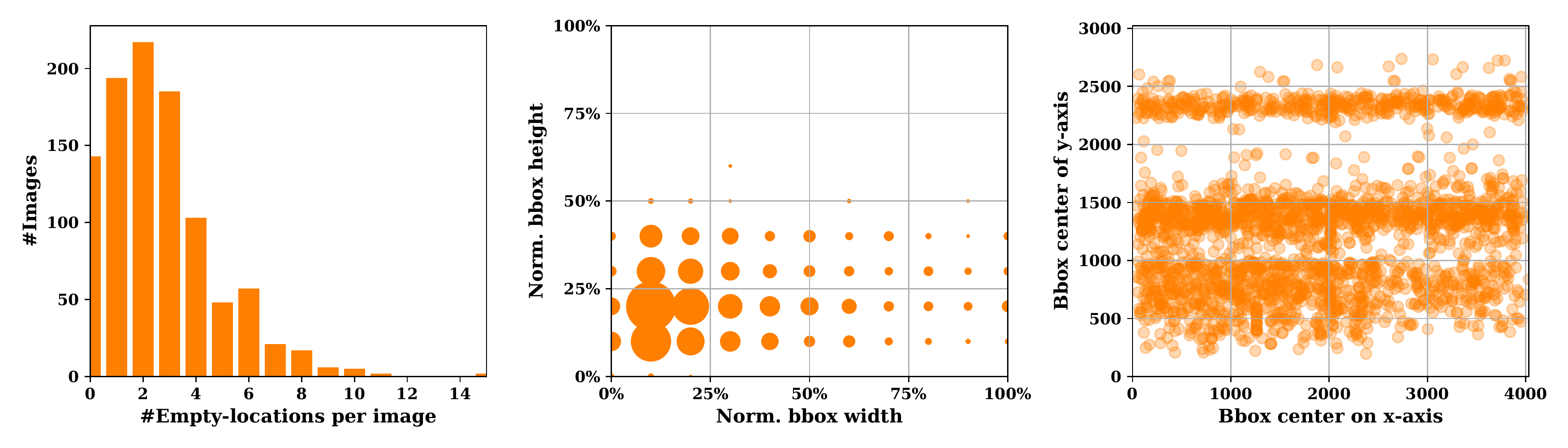}
  \caption{Histogram of empty-shelve count distribution in each image (left), histogram of bounding box size distribution (center), and bounding box center positions (right) in the dataset.}
  \Description{}
  \label{fig:data-dist}
\end{figure*}

\section{Data Annotation}

\textit{Empty shelf} is not a well-defined term; even two people working in the same team can have different and ambiguous interpretations of the same daily life concept. For example, how to determine the height of an empty location on the top shelf? How to handle empty locations above products? How to determine the bounding box corners for products placed in tilted positions or tilted images? Should a large empty location be marked as one or multiple? Where does an empty location end in the former case? Should a small gap between two products be marked as an empty location? To eliminate such ambiguities, we limit the annotations to well-defined concepts and established clear annotation guidelines as follows:

\begin{enumerate}
    \item Empty locations are created when a product is removed from its place by a customer. Since a product is 3D, we visualize an empty location as a 3D box to decide how to label the bounding box coordinates. Since bounding boxes are limited to 2D, we visualize the empty location as a 3D box and label the front face of the 3D box as shown in Figure~\ref{fig:labeling_3D}.
    \item Empty locations  have different heights or depths. When multiple products are placed on top of each other or horizontally next to each other from back to front, partially empty locations are created. In this work, we limit ourselves to completely empty locations - locations empty from front to back and top to bottom of the shelf (top, bottom, front and back faces of the visualized 3D cuboid do not touch any product).
    \item The empty location can be considered as multiple empty locations. When two adjacent products are removed, it creates a wider empty location. To avoid such confusion with labeling, we label a continuous empty location as a single empty location with one single bounding box.
    \item Sometimes there are small empty spaces between products because nothing can be placed there. To avoid such confusion, we ensure the width of the labeled empty location is at least half the size of the neighboring existing product.
    \item Shelf image contains several products along with empty locations. Since our focus is on empty locations, we do not label any product. This approach tremendously reduces the efforts required for data annotation (Figure~\ref{fig:labeling}).
    
\end{enumerate}

Using the above guidelines, we manually annotated the shelf images to create our dataset. Figure~\ref{fig:data-dist} illustrates different empty location and bounding box statistics. The left subplot in Figure~\ref{fig:data-dist} shows the histogram of empty location counts per image. The count of empty locations in an image ranges from 0 to 15. We included the images without any empty location as well so that the model can learn the representation of a completely filled-up shelf. The center subplot in Figure~\ref{fig:data-dist} shows the histogram of the normalized bounding box sizes; most of them are confined to smaller than 25\% of the image width and height, denoting empty locations created from removing just a few products. Note that the empty location height can not exceed the shelf partition; hence, the heights are confined to the lower portion of the graph. The right subplot of Figure~\ref{fig:data-dist} shows the bounding box center coordinates of the empty locations. Looking at this plot, we can see that they form three different rows; this represents three common levels of the shelves. After the data annotation phase, we created a dataset with train, validation and test sets containing 800, 100 and 100 samples, respectively.

\begin{figure*}[!t]
  \includegraphics[trim=10 0 10 10,clip, width=0.9\textwidth, keepaspectratio]{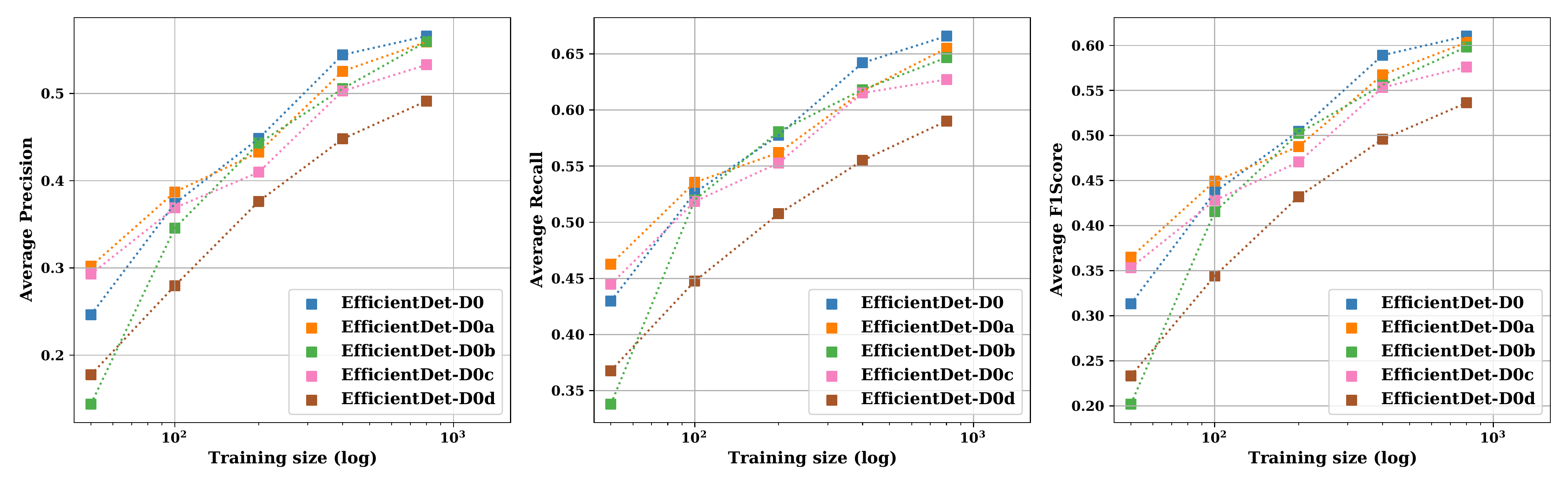}
  \caption{Learning curve for EfficientDet models with respect to training data size.}
  \Description{}
  \label{fig:learning_curve}
\end{figure*}

\section{Model Training}

In this section, we discuss how we trained different deep learning models for empty shelf detection. Since our goal is to build an efficient real-time empty shelf detection model, we focus on leveraging computationally-efficient deep neural network architectures. First, we train the EfficientDet~\cite{tan2020efficientdet} models since we could optimize their inference runtime for deployment on general-purpose CPUs, then we experiment with the other latest real-time object detection models from YOLOv5 family~\cite{yolov5}; the model training and inference codes are publicly available at their respective GitHub repositories.

\subsection{Learning Curve}

In this section, we answer the question - how many shelf images do we need to build a robust and accurate empty shelf detection model? To accomplish this, we follow an incremental approach to data collection by training EfficientDet models until the performance is saturated; we keep collecting data and adding to our training set as long as the learning curve keeps improving, keeping the validation and test set size fixed. 
For the deep neural network architecture, we leverage the EfficientDet-D0 along with its smaller variants, configurations are listed in Table~\ref{tab:efficientdet-d0s}. The model variants are created to analyze the learning capability of the neural network as dataset size is gradually increased. Since we start from a really small training set containing only 50 samples, hyperparameters are optimized for each model at each training dataset size using grid search over learning rate, learning rate patience, and batch size. To further utilize the capability of our models, we experiment using data augmentation borrowed from ScaledYOLOv4~\cite{wang2021scaled} with multiple augmentation settings from a combination of left-right flipping, rotations of the image, image translations, and random perspectives. We also experiment with different approaches for transfer learning from a pretrained EfficientDet-D0 model on MSCOCO - freeze the BiFPN, freeze the BiFPN+EfficientNet, and train all model layers. All models are trained using AdamW~\cite{loshchilov2017decoupled} optimization algorithm for 500 epochs using a patience of 30 for early stopping. The validation and test sets are fixed independent of the training set size. We select the models with the best performance on the validation set; since the precision and recall both are important for our application, we select models using the following metrics- mean average precision (mAP @ [IOU=0.50:0.95 | area=all | maxDets=100]), mean average recall (mAR @ [IOU=0.50:0.95 | area=all | maxDets=100]), and mean average F1-score (mAF) computed using previous two metrics.

\begin{table}[!t]
    \caption{Scaling configurations for EfficientDets used in Learning curve study.}
    \label{tab:efficientdet-d0s}
    \centering
    \footnotesize{
    \begin{tabular}{l|c|c|c|c|c}
    \toprule
    Model & Input & Backbone & \multicolumn{2}{c|}{BiFPN} & Box/class \\
    Name & size & Network & $\#$channels & $\#$layers & $\#$layers \\
    & $R_{input}$ & & $W_{bifpn}$ & $D_{bifpn}$ & $D_{class}$ \\ \hline
    \midrule
        D0 ($\phi=0$) & 512 & B0 & 64 & 3 & 3 \\
        D0a ($\phi=0$) & 512 & B0 & 64 & 1 & 1 \\
        D0b ($\phi=0$) & 512 & B0 & 32 & 3 & 3 \\
        D0c ($\phi=0$) & 512 & B0 & 32 & 1 & 1 \\
        D0d ($\phi=0$) & 256 & B0 & 32 & 1 & 1 \\
    \bottomrule
    \end{tabular}
    }
    \label{tab:my_label}
\end{table}

Figure~\ref{fig:learning_curve} demonstrates the learning curves of various model architectures for various performance metrics. We find almost a linear increase in model performance with a logarithmic increase in training set size.
The best models are generally ones leveraging data augmentation and slightly vary based on performance metric. For smaller training data, we observe that the smaller variants of EfficientDet-D0 perform better than the EfficientDet-D0 model. As the training size is increased, the model capacity comes into play; in fact, we observe that the performance, though similar for all variants, depends directly on the model size. The model size decreases in the following order - EfficientDet-D0a, EfficientDet-D0b, EFficientDet-D0c, and EfficientDet-D0d; the latter two have the same model architecture but differ in input resolution. Comparing these two models, we observe the performance lines are parallel to each other for all three metrics; higher input resolution provides more input features for the models to learn from and hence, perform better. Note that the model learning capabilities are not fully saturated with the largest training set containing 800 images in our dataset; we leave further exploration using more data collection for the future.

\begin{table}[!htb]
  \caption{Experimental results using different models}
  \label{tab:models}
  \footnotesize{
  \begin{tabular}{l|c|c|c|c|c|c|c|c}
    \toprule
    \multirow{2}{*}{Model} & Params & Input & \multicolumn{3}{|c|}{Validation} & \multicolumn{3}{|c}{Test} \\
    & (M) & Size & mAP & mAR & mAF & mAP & mAR & mAF \\ \hline
    \midrule
   
    EfficientDet-D0d & 3.65 & 256 & 49.1 & 59.0 & 53.6 & 48.4 & 57.6 & 52.8 \\ \hline
    EfficientDet-D0c & 3.65 & 512 & 53.3 & 62.7 & 57.6 & 55.1 & 63.1 & 58.8 \\ \hline
    EfficientDet-D0b & 3.68 & 512 & 56.0 & 64.3 & 59.8 & 54.0 & 63.0 & 58.2 \\ \hline
    EfficientDet-D0a & 3.73 & 512 & 55.9 & 65.6 & 60.3 & 55.6 & 63.6 & 59.3 \\ \hline
    EfficientDet-D0 & 3.83 & 512 & 56.6 & 66.3 & 61.0 & 55.3 & 64.0 & 59.3 \\ \hline
    EfficientDet-D1 & 6.55 & 640 & 57.9 & 66.4 & 61.9 & 57.0 & 63.8 & 60.2 \\ \hline
    EfficientDet-D2 & 8.01 & 768 & 49.0 & 59.7 & 53.8 & 50.4 & 60.7 & 55.0 \\ \hline
    \hline
    YOLOv5n & 1.76 & 640 & 66.2 & 76.7 & 71.1 & 63.8 & 74.0 & 68.5 \\ \hline
    YOLOv5n6 & 3.09 & 1280 & 68.3 & 78.9 & 73.2 & 66.9 & 76.3 & 71.3 \\ \hline
    YOLOv5s & 7.01 & 640 & 66.0 & 76.2 & 70.8 & 66.5 & 74.7 & 70.4 \\ \hline
    YOLOv5s6 & 12.31 & 1280 & 68.0 & 77.0 & 72.2 & 66.4 & 73.9 & 69.9 \\ \hline
    YOLOv5m & 20.85 & 640 & 68.9 & 75.4 & 72.0 & 65.3 & 72.8 & 68.8 \\ \hline
    YOLOv5m6 & 35.25 & 1280 & 67.3 & 75.4 & 71.1 & 64.3 & 72.9 & 68.3 \\ \hline
    YOLOv5l & 46.11 & 640 & 67.7 & 74.8 & 71.1 & 66.0 & 73.3 & 69.5 \\ \hline
    YOLOv5l6 & 76.12 & 1280 & 69.2 & 77.1 & 72.9 & 65.9 & 73.7 & 69.6 \\ \hline
    YOLOv5x & 86.17 & 640 & 66.5 & 76.3 & 71.0 & 66.9 & 74.4 & 70.5 \\ \hline
    YOLOv5x6 & 139.97 & 1280 & 67.7 & 75.7 & 71.5 & 65.8 & 73.2 & 69.3 \\ \hline

  \bottomrule
\end{tabular}
}
\end{table}

\subsection{Performance Evaluation}

Once the learning curves started to plateau, we experimented with the latest state-of-the-art computationally-efficient deep learning model architectures. We selected the EfficientDets and the YOLOv5 model family for their best real-time performance. For the YOLOv5 models, we use the default parameters along with a batch size of 4 and trained for 90 epochs after initializing the model parameters from pretrained models on MSCOCO. For the EfficientDet models, we perform the hyperparameter search as mentioned before. Since both average precision and average recall are important for our empty shelf detection, we use mean average F1-score (mAF computed using AP and AR) to compare model performance.

Table~\ref{tab:models} demonstrates the performance of different models on the validation set and test set; all the models are trained on the training set containing 800 images, and we report the results of the models with the best performance on the validation set. The EfficientDet models are sorted in the increasing order of model parameters. EfficientDet-D0c significantly outperforms the EfficientDet-D0d model as it can take advantage of $4\times$ more features present in the input image. Reducing the number of BiFPN repetitions does not seem to impact the model performance; EfficientDet-D0a performs similar to EfficientDet-D0. When we reduce the number of channels in the BiFPN layers, the model performance has an observable loss. Comparing the performance of larger EfficientDet models, we observe a slight increase in performance going from EfficientDet-D0 to EfficientDet-D1, but the performance suffers when increasing the model architecture size to EfficientDet-D2. We conjecture the loss in performance with an increase in model size results from the limited size of our dataset. The best EfficientDet model from this analysis is the EfficientDet-D1 with an average precision of 57.0\%, average recall of 63.8\% and average F1-score of 60.2\%. This indicates that we have saturated the learning capability of EfficientDet models for our modeling problem; hence, we decided not to proceed with training larger EfficientDet models.

\begin{figure*}[!tb]
  \includegraphics[width=0.99\textwidth, keepaspectratio]{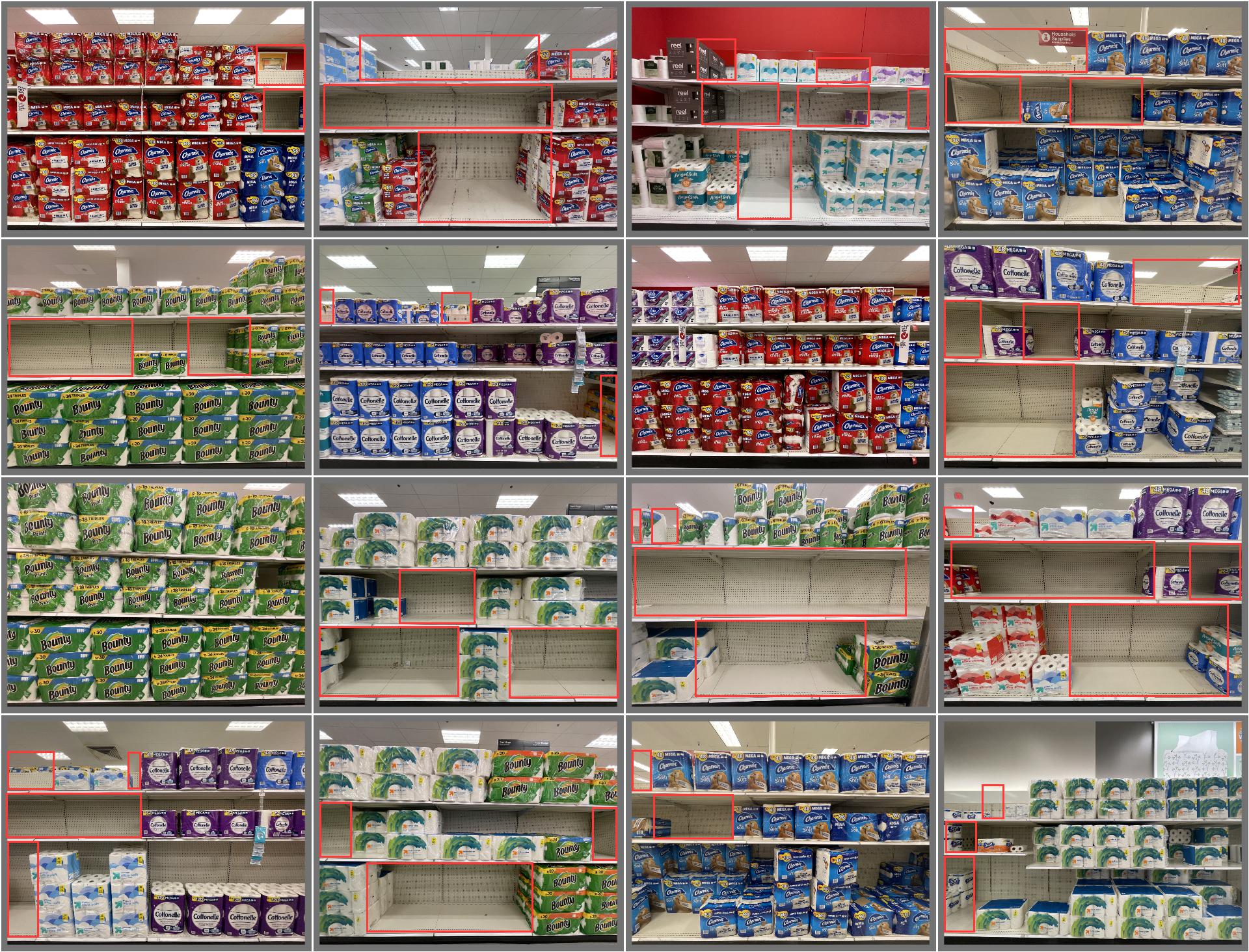}
  \caption{Model predictions using YOLOv5n6 for empty shelf detection.}
  \Description{}
  \label{fig:prediction}
\end{figure*}

Next, we analyze the performance of YOLOv5 model architectures on our dataset for empty shelf detection. The YOLOv5 model family is composed of model architectures of varying sizes - nano (YOLOv5n), small (YOLOv5s), medium (YOLOv5m), large (YOLOv5l) and extra large (YOLOv5x). They are generally trained using an input resolution of 640; however, the YOLOv5-6 variants have around 1.5x parameters compared to corresponding YOLOv5 models and they are trained using an input resolution of 1280. First, looking at the models trained using input resolution of 640, we observe that different models have comparable performance; YOLOv5n has the lowest average precision of 63.8\%; YOLOv5s has better precision than YOLOv5m and YOLOv5l; YOLOv5n, YOLOv5s and YOLOv5x has high recall. Next, if we look at the impact of increasing the input resolution to 1280, we observe a clear impact for YOLOv5n, the model performance increases significantly from an average F1-score of 68.5\% to 71.3\%. For other models, there is a slight decrease in performance.
Considering the best average F1-Score on the test set, the YOLOv5 model with the best performance is YOLOv5n6 (Figure~\ref{fig:prediction} illustrates predictions using YOLOv5n6).

This clearly demonstrates that larger model architectures may not always perform best, and hence, the model selection should depend on the dataset size and complexity of the given modeling task. Note that YOLOv5 performances are significantly (>10\%) higher than the EfficientDet models. This illustrates the importance of correctly labeling data to create high-quality dataset for modeling and the importance of proper model selection based on the dataset and the complexity of the given machine learning task.

\section{Inference Optimization}
There exists different performance trade-offs when deploying a model for in-production inference. Beyond the model metrics, we have to also consider the available computing resources and the network bandwidth in the deployment environment. In this section, we do analyze latency and throughput of the models using different inference run-time optimizations to determine which model is optimal for deployment on the GPUs in the data centers and the CPUs available in the retail stores. 
Table~\ref{tab:infer} illustrates the latency and throughput using different quantization and run-time optimization frameworks including PyTorch (model training framework) for both NVIDIA A100 GPU and Intel Xeon Gold CPU.

\begin{table*}[!h]
  \caption{Inference run-time optimization results on different devices.}
  \label{tab:infer}
  \footnotesize{
  \begin{tabular}{l|c|c|c|c|c|c|c|c|c|c|c|c}
    \toprule
    \multirow{5}{*}{Model} & \multirow{5}{*}{Params} & \multicolumn{4}{|c|}{A100 GPU} & \multicolumn{7}{|c}{Intel Xeon Gold 6148 @ 2.40 GHZ} \\
    \cline{3-13}
    & & \multicolumn{4}{|c|}{PyTorch} & \multicolumn{2}{|c|}{PyTorch} & \multicolumn{1}{|c|}{ONNX} & \multicolumn{4}{|c}{OpenVINO} \\
    \cline{3-13}
    
    & & \multicolumn{2}{|c|}{FP32} & \multicolumn{2}{|c|}{FP16} & \multicolumn{2}{|c|}{FP32} & FP32 & \multicolumn{2}{|c|}{FP32} & \multicolumn{2}{|c}{FP16} \\ \cline{3-13}
    & (M) & Lat. & TP [BS=32] & Lat. & TP [BS=32] & Lat. & TP [BS=32] & Lat. & Lat. & TP [BS=32] & Lat. & TP [BS=32] \\
    & & ms & images/sec & ms & images/sec & ms & images/sec & ms & ms & images/sec & ms & images/sec \\ \hline 
    
    \midrule
    YOLOv5n & 1.76 & 10.1 & 860.4 & 11.4\textit{(1.1x)} & 838.1\textit{(1.0x)} & 54.2 & 3.9 & 41.0\textit{(0.8x)} & 14.6\textit{(0.3x)} & 67.2\textit{(17.1x)} & 19.2\textit{(0.4x)} & 67.8\textit{(17.2x)} \\ \hline
    YOLOv5n6 & 3.09 & 13.7 & 400.3 & 15.3\textit{(1.1x)} & 375.2\textit{(0.9x)} & 151.5 & 2.9 & 107.5\textit{(0.7x)} & 57.1\textit{(0.4x)} & 16.4\textit{(5.7x)} & 53.4\textit{(0.4x)} & 14.5\textit{(5.0x)} \\ \hline
    YOLOv5s & 7.01 & 11.7 & 671.9 & 11.8\textit{(1.0x)} & 675.2\textit{(1.0x)} & 112.6 & 3.8 & 63.6\textit{(0.6x)} & 39.7\textit{(0.4x)} & 38.9\textit{(10.3x)} & 39.8\textit{(0.4x)} & 39.9\textit{(10.6x)} \\ \hline
    YOLOv5s6 & 12.31 & 15.6 & 258.6 & 15.9\textit{(1.0x)} & 235.5\textit{(0.9x)} & 394.1 & 2.4 & 161.9\textit{(0.4x)} & 84.3\textit{(0.2x)} & 9.5\textit{(3.9x)} & 87.1\textit{(0.2x)} & 9.1\textit{(3.8x)} \\ \hline
    YOLOv5m & 20.85 & 14.1 & 427.7 & 14.9\textit{(1.1x)} & 518.1\textit{(1.2x)} & 221.6 & 3.1 & 77.9\textit{(0.4x)} & 53.5\textit{(0.2x)} & 21.2\textit{(6.8x)} & 56.4\textit{(0.3x)} & 21.7\textit{(7.0x)} \\ \hline
    YOLOv5m6 & 35.25 & 18.2 & 137.1 & 19.5\textit{(1.1x)} & 176.2\textit{(1.3x)} & 841.9 & 1.9 & 287.0\textit{(0.3x)} & 154.2\textit{(0.2x)} & 5.1\textit{(2.7x)} & 147.6\textit{(0.2x)} & 5.0\textit{(2.7x)} \\ \hline
    YOLOv5l & 46.11 & 17.2 & 347.7 & 18.0\textit{(1.0x)} & 402.9\textit{(1.2x)} & 391.3 & 2.7 & 131.5\textit{(0.3x)} & 90.8\textit{(0.2x)} & 10.8\textit{(4.0x)} & 90.2\textit{(0.2x)} & 11.6\textit{(4.3x)} \\ \hline
    YOLOv5l6 & 76.12 & 21.9 & 96.4 & 23.4\textit{(1.1x)} & 129.0\textit{(1.3x)} & 1462.7 & 1.3 & 466.4\textit{(0.3x)} & 227.2\textit{(0.2x)} & 3.0\textit{(2.3x)} & 229.2\textit{(0.2x)} & 3.0\textit{(2.3x)} \\ \hline
    YOLOv5x & 86.17 & 19.6 & 218.7 & 20.7\textit{(1.1x)} & 287.4\textit{(1.3x)} & 632.0 & 2.2 & 194.8\textit{(0.3x)} & 141.8\textit{(0.2x)} & 6.6\textit{(3.0x)} & 133.4\textit{(0.2x)} & 6.5\textit{(3.0x)} \\ \hline
    YOLOv5x6 & 139.97 & 33.2 & 64.1 & 25.9\textit{(0.8x)} & 84.7\textit{(1.3x)} & 2578.2 & 0.9 & 730.9\textit{(0.3x)} & 370.4\textit{(0.1x)} & 2.0\textit{(2.2x)} & 353.3\textit{(0.1x)} & 2.0\textit{(2.2x)} \\ \hline
  \bottomrule
\end{tabular}}
\end{table*}

\begin{figure*}[!tb]
  \includegraphics[trim=70 0 70 20,clip,width=\textwidth, keepaspectratio]{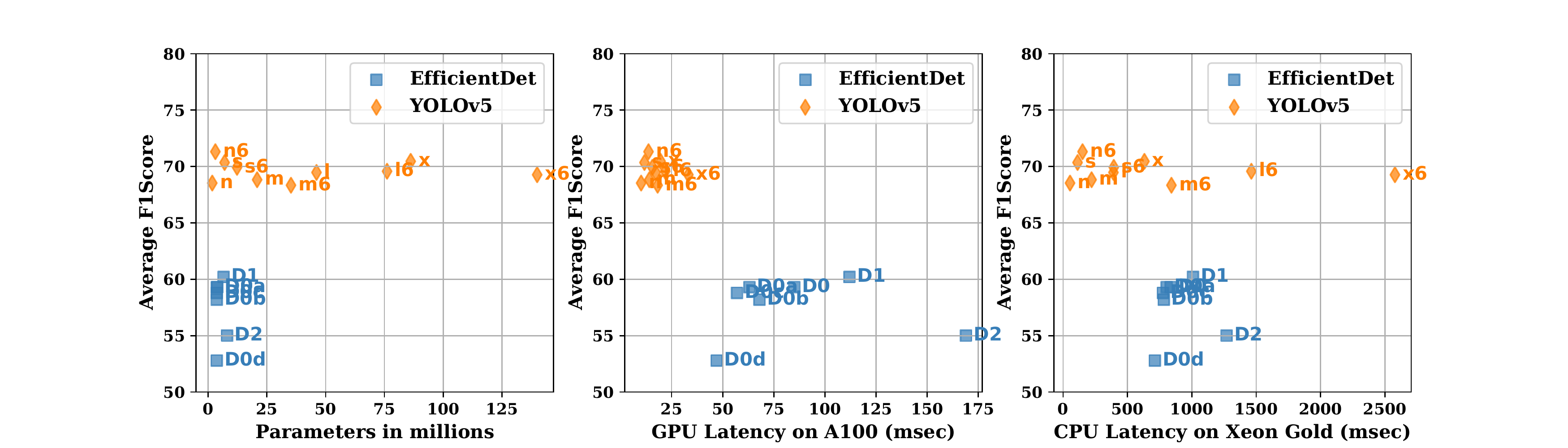}
  \caption{Inference latencies on NVIDIA A100 GPU and Intel Xeon Gold CPU using PyTorch and FP32.}
  \Description{}
  \label{fig:inference_lat}
\end{figure*}

\subsection{Impact of Model Parameters}
Generally, the model performance should increase with an increase in model parameters since that means an increase in learning capability, provided the dataset is large enough. Since our dataset only contains 800 samples in the training set, we do not expect the performance to improve with an increase in model parameters. Figure~\ref{fig:inference_lat} illustrates model performance trade-off against model parameters (left subplot), GPU latency on A100 (center subplot) and CPU latency on Xeon Gold (right subplot). We can observe that performance improved with an increase in model parameters for EfficientDet-D0 variants. However, this trend is not observed when moving from EfficientDet-D1 to EfficientDet-D2 and for the YOLOv5 models. This implies that the appropriate model size is really dependent upon the quality, size and complexity of the dataset. For a small and high-quality dataset, we do not really need huge models to achieve great performance (as you can see YOLOv5n6 has the best performance, although it is the second smallest model in our whole study).

The number of model parameters is also an important criterion to consider before deployment since that determines the model size, and hence, the required amount of memory to run the model. Table~\ref{tab:models} also enlists the parameter counts for different models. The model with the best performance is YOLOv5n6, having 3.09M parameters, which takes a relatively small amount of memory to run.

\subsection{Inference Optimization on A100}

From the accuracy vs. GPU latency curve in Figure~\ref{fig:inference_lat}, we observe that the YOLOv5 models have small GPU latencies between 10 ms to 20 ms, the YOLOv5-6 models having slightly higher latencies between 14 ms to 33 ms; all YOLOv5 models have latencies below 22ms except for YOLOv5x6. In comparison, the EfficientDets have significantly higher GPU latencies above 50ms in almost all cases, making them at least 5 times slower. Analysing the GPU throughput in Table~\ref{tab:infer}, the YOLOv5n model can process around 860 images/sec on an A100 GPU using a batch size of 32, while having comparable performance to YOLOv5n6, which can process only around 400 images/sec on an A100 GPU. Note that YOLOv5n has the smallest GPU latency of 10.1 ms, compared to YOLOv5n6 having a latency of 13.7 ms on A100 -- latencies are measured at a batch size of 1. The YOLOv5s model with F1-score of 70.4 can process around 672 images/s on an A100 GPU.

The YOLOv5 models also support quantization to FP16 on GPUs as shown in Table~\ref{tab:infer}. 
For the smaller YOLOv5 models, we do not see any improvement from using FP16; in fact, we observe a slight increase in latency for the smaller models. However, for the larger models starting from YOLOv5m, we do observe an increase in throughput, though the latency remains similar. For the largest model YOLOv5x6, the latency reduces to 0.8$\times$ and throughput increases to $1.3x$; increase in throughput is similar for other larger models. This demonstrates the importance of analyzing the performance tradeoffs -- accuracy vs latency and throughput, as well as utilizing the available inference optimizations to satisfy the performance requirements before proceeding with deployment on GPUs.

\subsection{Inference Optimization on Intel Xeon Gold}

The performance vs. latency trade-off looks significantly different when we look at Intel CPUs. Here, we used the Intel Xeon Gold 6148 CPU @ 2.40GHz with 20 physical cores for our CPU inference analysis. Again, the YOLOv5 models appear to be the clear winner when compared against the EfficientDet models. YOLOv5n has the lowest CPU latency of 54 ms and YOLOv5n6 has a CPU latency of 151 ms in PyTorch. If we look at the results using a batch size of 32 in Table~\ref{tab:infer}, we observe that YOLOv5 models do not perform well with an increase in batch size. For a batch size of 32, YOLOv5n has only a throughput of 3.9 images/s; although it can process almost 18 images/sec using a batch size of 1. We observe similar characteristics for the first five smallest YOLOv5 models. Note that here were are using the models with PyTorch runtime. From these analyses and observations, we conclude that smaller YOLOv5 models with PyTorch runtime should be used with a batch size of 1 on this CPU since it has high throughput @ batch size of 1. We conjecture this happens because the smaller models can almost fit into the L3 cache (27.5 MB) of the CPU and increasing batch size might be increasing the rate of page thrashing; we plan to investigate more in the future.

Furthermore, we leverage the ONNX and OpenVINO runtime optimization frameworks for improving runtime performance of our YOLOv5 models on Intel CPU (Table~\ref{tab:infer}). Using ONNX runtime, we observe that the latency decreases significantly for the larger models starting with YOLOv5s6; for smaller models such as YOLOv5n, the latency decreases slightly (0.8$\times$). We further optimize the ONNX models using OpenVINO, both FP32 and FP16. OpenVINO significantly reduces the model latencies, even for smaller models. YOLOv5n latency reduces from 54 ms to 14.6 ms when converting from PyTorch to OpenVINO; the latency reduces by 10$\times$ for the largest model - YOLOv5x6. We observe significant improvement in inference throughput as well - the throughput increases from 3.9 images/s using PyTorch, to 67.2 images/s using OpenVINO (FP32). For the largest model - YOLOv5x6, the throughput increases by a factor of 2.2$\times$ when compared against PyTorch framework. We do not observe any clear benefit from using FP16 for the OpenVINO framework; the latency and throughput of models using OpenVINO framework are similar when using FP32 and FP16. An interesting observation is that the latency improvement increases with increase in model parameters, while the throughput improvement decreases with increase in model parameters, as we move from PyTorch to OpenVINO framework. The significant improvements -- upto 10$\times$ in latency for YOLOv5x6 and upto 17.1$\times$ in throughput for YOLOv5n, demonstrates the critical importance of leveraging run-time optimization for model inference before proceeding with deployment.

\section{Deployment}
The final stage in a machine learning pipeline is the proper deployment of the optimized model for real-time inference. The actual deployment is based on the customer requirements and the deployment infrastructure. As seen in the previous section, an optimized YOLOv5 model can process more than 860 images/s on an A00 GPU. However, installing a powerful GPU, such as an A100 GPU, in thousands of retail stores is not ideal. One might suggest processing the shelf images from retail stores using GPUs in data centers; however, a critical issue with that would be transmitting the shelf images continuously from retail stores to the data centers over internet; these images are generally huge in size, and the network bandwidths in retail stores are often limited, deeming such an approach infeasible. A feasible approach for deployment would be to deploy the models inside the stores themselves on a traditional CPU server. Note that existing servers inside stores have limited computing capabilities, having thousands of jobs to run; quickly upgrading the computing infrastructure across thousands of retail stores is also not practical.

The optimized models have been deployed in our production environment. To analyze the deployment run-time,  we analyzed an optimized model deployed on an Intel Xeon Gold in our data center; it takes shelf images from an incoming Kafka stream as input and sends back the empty location pixel coordinates as the output using a Kafka stream. Here are some interesting findings from our deployment analysis:

\begin{enumerate}
    \item Different steps beyond the model can significantly impact the inference time. For example, sometimes, image decoding and preprocessing can take significant time; it is critical to optimize these computations using parallelization available in PyTorch and other frameworks.
    
    \item The maximum throughput of a deployed model can depend on many factors, batch size being one of them. It is crucial to analyze the trade-offs of these factors before selecting the batch size for model processing.
    
    \item Determining the appropriate memory requirement for model deployment is critical; it can significantly impact the model performance (both latency and throughput).
    
    \item Inference run-time optimizations can significantly improve performance. For example, we were able to increase the throughput of YOLOv5n by 17$\times$ and EfficientDets by 2$\times$ using OpenVINO before deployment.
    
    \item Models can only perform well on incoming data with a similar representation and distribution to the training data. It is important to keep track of data drift and retrain the model when necessary.
    
\end{enumerate}

\section{Significance and Impact}
OSA has become an critical indicator of the customer service output for a retailer performance~\cite{moorthy2015applying}.
A study by Corsten and Gruen~\cite{corsten2003desperately} indicated that OOS situations mostly happens at the store level, mainly due to issues in product ordering and replenishment practices in the retail stores. 
Currently, there exist several solutions to address the issue of OOS products. While the manual and technology-based (such as RFID) solutions are expensive to integrate and maintain, the image processing and machine learning based solutions have low accuracy and do not work well in real retail environment, and deep learning based solutions are limited by the availability of well-annotated datasets. In this work, we presented a high quality dataset and build an efficient high-accuracy deep learning model for real-time empty-shelf detection in retail stores that can be deployed on store computers as well as in the data centers. We believe this work not only takes an important step in solving the OOS issue, but also present a well-defined machine learning pipeline for building real-world applications using AI.

\section{Conclusion and Future Works}
In this paper, we presented an elegant approach to design and build an efficient end-to-end real-time deep learning solution for empty-shelf detection. We elaborately discussed each machine learning pipeline stage involved in building this solution. We collected a small dataset following well-defined guidelines, annotated the data using clear rules and incrementally kept increasing our dataset size until the learning curve started saturating. We experimented with several state-of-the-art object detection models and performed thorough inference analysis, along with inference run-time optimizations supported by the respective models. We briefly discussed the deployment issues and shared our findings. We believe this work would be valuable not only to the retail industry but also others interested in building their own end-to-end efficient real-time deep learning solutions from scratch. We plan to pursue training with the other state-of-the-art models, inference time optimizations and deployment in various environments in the future.

\bibliographystyle{unsrt}
\bibliography{target-shelves}

\begin{thebibliography}{10}

\bibitem{fisher2000rocket}
Marshall~L Fisher, Ananth Raman, and Anna~Sheen McClelland.
\newblock Rocket science retailing is almost here-are you ready?
\newblock {\em Harvard Business Review}, 78(4):115--123, 2000.

\bibitem{anderson2006measuring}
Eric~T Anderson, Gavan~J Fitzsimons, and Duncan Simester.
\newblock Measuring and mitigating the costs of stockouts.
\newblock {\em Management science}, 52(11):1751--1763, 2006.

\bibitem{gruen2007comprehensive}
Thomas~W Gruen, Daniel~S Corsten, et~al.
\newblock A comprehensive guide to retail out-of-stock reduction in the
  fast-moving consumer goods industry.
\newblock 2007.

\bibitem{spielmaker2012shelf}
Kristie Spielmaker.
\newblock On shelf availability: A literature review \& conceptual framework.
\newblock 2012.

\bibitem{musalem2010structural}
Andr{\'e}s Musalem, Marcelo Olivares, Eric~T Bradlow, Christian Terwiesch, and
  Daniel Corsten.
\newblock Structural estimation of the effect of out-of-stocks.
\newblock {\em Management Science}, 56(7):1180--1197, 2010.

\bibitem{gruen2002retail}
Thomas~W Gruen, Daniel~S Corsten, and Sundar Bharadwaj.
\newblock {\em Retail out-of-stocks: A worldwide examination of extent, causes
  and consumer responses}.
\newblock Grocery Manufacturers of America Washington, DC, 2002.

\bibitem{mitchell2012improving}
Andrew Mitchell.
\newblock Improving on-shelf availability.
\newblock {\em White paper, Symphony IRI Group}, 2012.

\bibitem{chao2007determining}
Chia-Chen Chao, Jiann-Min Yang, and Wen-Yuan Jen.
\newblock Determining technology trends and forecasts of rfid by a historical
  review and bibliometric analysis from 1991 to 2005.
\newblock {\em Technovation}, 27(5):268--279, 2007.

\bibitem{milella20213d}
Annalisa Milella, Roberto Marani, Antonio Petitti, Grazia Cicirelli, and
  Tiziana D’Orazio.
\newblock 3d vision-based shelf monitoring system for intelligent retail.
\newblock In {\em International Conference on Pattern Recognition}, pages
  447--459. Springer, 2021.

\bibitem{moorthy2015applying}
Rahul Moorthy, Swikriti Behera, Saurav Verma, Shreyas Bhargave, and Prasad
  Ramanathan.
\newblock Applying image processing for detecting on-shelf availability and
  product positioning in retail stores.
\newblock In {\em Proceedings of the Third International Symposium on Women in
  Computing and Informatics}, pages 451--457, 2015.

\bibitem{michael2005pros}
Katina Michael and Luke McCathie.
\newblock The pros and cons of rfid in supply chain management.
\newblock In {\em International Conference on Mobile Business (ICMB'05)}, pages
  623--629. Ieee, 2005.

\bibitem{rosado2016supervised}
Lu{\'\i}s Rosado, Jo{\~a}o Gon{\c{c}}alves, Jo{\~a}o Costa, David Ribeiro, and
  Filipe Soares.
\newblock Supervised learning for out-of-stock detection in panoramas of retail
  shelves.
\newblock In {\em 2016 IEEE International Conference on Imaging Systems and
  Techniques (IST)}, pages 406--411. IEEE, 2016.

\bibitem{muthugnanambika2018automated}
M~Muthugnanambika, T~Bagyammal, Latha Parameswaran, and Karthikeyan Vaiapury.
\newblock An automated vision based change detection method for planogram
  compliance in retail stores.
\newblock In {\em Computational Vision and Bio Inspired Computing}, pages
  399--411. Springer, 2018.

\bibitem{milella2020towards}
Annalisa Milella, Antonio Petitti, Roberto Marani, Grazia Cicirelli, and
  Tiziana D’orazio.
\newblock Towards intelligent retail: Automated on-shelf availability
  estimation using a depth camera.
\newblock {\em IEEE Access}, 8:19353--19363, 2020.

\bibitem{priyanwada2020benchmark}
H.A.M Priyanwada, K.A.D~Dilanka Madhushan, Chethana Liyanapathirana, and Lakmal
  Rupasinghe.
\newblock Vision based intelligent shelf-management system.
\newblock In {\em 2021 6th International Conference on Information Technology
  Research (ICITR)}, pages 1--6, 2021.

\bibitem{higa2019robust}
Kyota Higa and Kota Iwamoto.
\newblock Robust shelf monitoring using supervised learning for improving
  on-shelf availability in retail stores.
\newblock {\em Sensors}, 19(12):2722, 2019.

\bibitem{chen2019out}
Jun Chen, Shu-Lin Wang, and Hong-Li Lin.
\newblock Out-of-stock detection based on deep learning.
\newblock In {\em International Conference on Intelligent Computing}, pages
  228--237. Springer, 2019.

\bibitem{rong2020solution}
Tianze Rong, Yanjia Zhu, Hongxiang Cai, and Yichao Xiong.
\newblock A solution to product detection in densely packed scenes.
\newblock {\em arXiv preprint arXiv:2007.11946}, 2020.

\bibitem{yilmazer2021shelf}
Ramiz Yilmazer and Derya Birant.
\newblock Shelf auditing based on image classification using semi-supervised
  deep learning to increase on-shelf availability in grocery stores.
\newblock {\em Sensors}, 21(2):327, 2021.

\bibitem{goldman2019precise}
Eran Goldman, Roei Herzig, Aviv Eisenschtat, Jacob Goldberger, and Tal Hassner.
\newblock Precise detection in densely packed scenes.
\newblock In {\em Proceedings of the IEEE/CVF Conference on Computer Vision and
  Pattern Recognition}, pages 5227--5236, 2019.

\bibitem{varadarajan2020benchmark}
Srikrishna Varadarajan, Sonaal Kant, and Muktabh~Mayank Srivastava.
\newblock Benchmark for generic product detection: a low data baseline for
  dense object detection.
\newblock In {\em International Conference on Image Analysis and Recognition},
  pages 30--41. Springer, 2020.

\bibitem{wei2020deep}
Yuchen Wei, Son Tran, Shuxiang Xu, Byeong Kang, and Matthew Springer.
\newblock Deep learning for retail product recognition: Challenges and
  techniques.
\newblock {\em Computational intelligence and neuroscience}, 2020, 2020.

\bibitem{karpathy2017software}
A~Karpathy.
\newblock Software 2.0 a medium corporation, programming.
\newblock {\em Dostupno na: https://medium. com/@
  karpathy/software-2-0-a64152b37c35}, 2017.

\bibitem{makinen2021needs}
Sasu M{\"a}kinen, Henrik Skogstr{\"o}m, Eero Laaksonen, and Tommi Mikkonen.
\newblock Who needs mlops: What data scientists seek to accomplish and how can
  mlops help?
\newblock {\em arXiv preprint arXiv:2103.08942}, 2021.

\bibitem{renggli2021data}
Cedric Renggli, Luka Rimanic, Nezihe~Merve G{\"u}rel, Bojan Karla{\v{s}},
  Wentao Wu, and Ce~Zhang.
\newblock A data quality-driven view of mlops.
\newblock {\em arXiv preprint arXiv:2102.07750}, 2021.

\bibitem{batini2009methodologies}
Carlo Batini, Cinzia Cappiello, Chiara Francalanci, and Andrea Maurino.
\newblock Methodologies for data quality assessment and improvement.
\newblock {\em ACM computing surveys (CSUR)}, 41(3):1--52, 2009.

\bibitem{webmarket}
Kaggle.
\newblock Webmarket dataset.
\newblock \url{https://www.kaggle.com/manikchitralwar/webmarket-dataset}, 2022.

\bibitem{corsten2003desperately}
Daniel Corsten and Thomas Gruen.
\newblock Desperately seeking shelf availability: an examination of the extent,
  the causes, and the efforts to address retail out-of-stocks.
\newblock {\em International Journal of Retail \& Distribution Management},
  2003.

\bibitem{hausruckinger2006approaches}
Gerhard Hausruckinger.
\newblock Approaches to measuring on-shelf availability at the point of sale.
\newblock {\em Preuzeto sa http://ecr-all. org/content/ecropedia\_element.
  php}, 2006.

\bibitem{campo2000towards}
Katia Campo, Els Gijsbrechts, and Patricia Nisol.
\newblock Towards understanding consumer response to stock-outs.
\newblock {\em Journal of Retailing}, 76(2):219--242, 2000.

\bibitem{tan2020efficientdet}
Mingxing Tan, Ruoming Pang, and Quoc~V Le.
\newblock Efficientdet: Scalable and efficient object detection.
\newblock In {\em Proceedings of the IEEE/CVF conference on computer vision and
  pattern recognition}, pages 10781--10790, 2020.

\bibitem{yolov5}
Ultralytics Inc.
\newblock Github - ultralytics/yolov5.
\newblock \url{https://github.com/ultralytics/yolov5}, 2022.

\bibitem{wang2021scaled}
Chien-Yao Wang, Alexey Bochkovskiy, and Hong-Yuan~Mark Liao.
\newblock Scaled-yolov4: Scaling cross stage partial network.
\newblock In {\em Proceedings of the IEEE/CVF Conference on Computer Vision and
  Pattern Recognition}, pages 13029--13038, 2021.

\bibitem{loshchilov2017decoupled}
Ilya Loshchilov and Frank Hutter.
\newblock Decoupled weight decay regularization.
\newblock {\em arXiv preprint arXiv:1711.05101}, 2017.

\bibitem{ehrenthal2012service}
Joachim~CF Ehrenthal.
\newblock {\em A Service-Dominant Logic view of retail on-shelf availability}.
\newblock PhD thesis, Rohner+ Spiller AG, 2012.

\bibitem{lambert2006fundamentals}
D~Lambert, D~Grant, J~Stock, and L~Ellram.
\newblock Fundamentals of logistics management, european edition.
\newblock {\em Maidenhead, Berkshire: McGraw-Hill}, 2006.

\end{thebibliography}

\end{document}